\documentclass{article}
 \pdfoutput=1

\usepackage{spconf,amsmath,amssymb,graphicx}
\usepackage{enumitem}
\setlist{nosep, leftmargin=14pt}
\usepackage{booktabs}
\usepackage[table,xcdraw]{xcolor}
\usepackage{comment}
\usepackage{tablefootnote}
\usepackage[hidelinks]{hyperref}

\title{An automated pipeline to create an atlas of \emph{in situ} hybridization \\ gene expression data in the adult marmoset brain}

\name{%
\begin{tabular}{@{}c@{}}
Charissa Poon $^{\star \ast}$  \qquad 
Muhammad Febrian Rachmadi$^{\star}$ \qquad 
Michal Byra$^{\star \dagger}$ \\ 
Matthias Schlachter$^{\star}$ \qquad 
Binbin Xu$^{\star \dagger\dagger}$ \qquad 
Tomomi Shimogori$^{\ddagger}$ \\
Henrik Skibbe$^{\star}$
\end{tabular}}

\address{$^{\star}$ Brain Image Analysis Unit, RIKEN Center for Brain Science, 
Wako, Japan \\
$^{\dagger}$ Institute of Fundamental Technological Research, Polish Academy of 
Sciences, Warsaw, Poland \\ 
$^{\dagger\dagger}$ EuroMov Digital Health in Motion, Univ Montpellier, IMT Mines Alès, Alès, France \\ 
$^{\ddagger}$ Lab for Molecular Mechanisms of Brain Development, RIKEN 
Center for Brain Science, Wako, Japan \\ \\
$^{\ast}$ \small Corresponding author: charissa.poon at riken.jp} % or add as a footnote

\begin{document}

\maketitle

\begin{abstract}
We present the first automated pipeline to create an atlas of \emph{in situ} hybridization gene expression in the adult marmoset brain in the same stereotaxic space. The pipeline consists of segmentation of gene expression from microscopy images and registration of images to a standard space. Automation of this pipeline is necessary to analyze the large volume of data in the genome-wide whole-brain dataset, and to process images that have varying intensity profiles and expression patterns with minimal human bias. To reduce the number of labelled images required for training, we develop a semi-supervised segmentation model. We further develop an iterative algorithm to register images to a standard space, enabling comparative analysis between genes and concurrent visualization with other datasets, thereby facilitating a more holistic understanding of primate brain structure and function.
\end{abstract}
\section{Introduction}
\label{sec:intro}

Characterization of gene expression in the brain is necessary to understand brain structure and function. Cellular diversity in the brain points at the need to characterize gene expression at single-cell resolution. Gene expression brain atlases in lower-order model organisms have led to better understanding of anatomical structures and cell types based on spatial expression patterns of genes. However, interspecies differences limits the extrapolation of findings to the human brain. The common marmoset (\emph{Callithrix jacchus}) exhibits human-like social traits, a fast reproductive cycle, and has proven to be amenable to genetic manipulation, characteristics that make it a candidate model organism for primate research. 

The \href{gene-atlas.brainminds.jp/}{Marmoset Gene Atlas}, created by the Brain/MINDS project in Japan, is an \emph{in situ} hybridization (ISH) database of gene expression in the neonate and adult marmoset brain \cite{Shimogori2018,Okano2016brain}. Characterization of neonate marmoset ISH gene expression images led to the discovery of regional- and species-specific patterns of gene expression in the developing marmoset brain \cite{Kita2021}. However, like other existing atlases \cite{Lein2006}, segmentation of ISH gene expression was conducted manually \cite{Kita2021}. Manual methods are susceptible to human bias and error and not feasible for characterizing gene expression on a whole-brain, genome-wide, multi-age level. Furthermore, existing marmoset brain atlases lack transcriptomic data such as the ISH dataset (e.g. \cite{Majka2020,Tian2022,Skibbe2022}).

\begin{figure}[htb]
\centering
\centerline{\includegraphics[width=8.5cm]{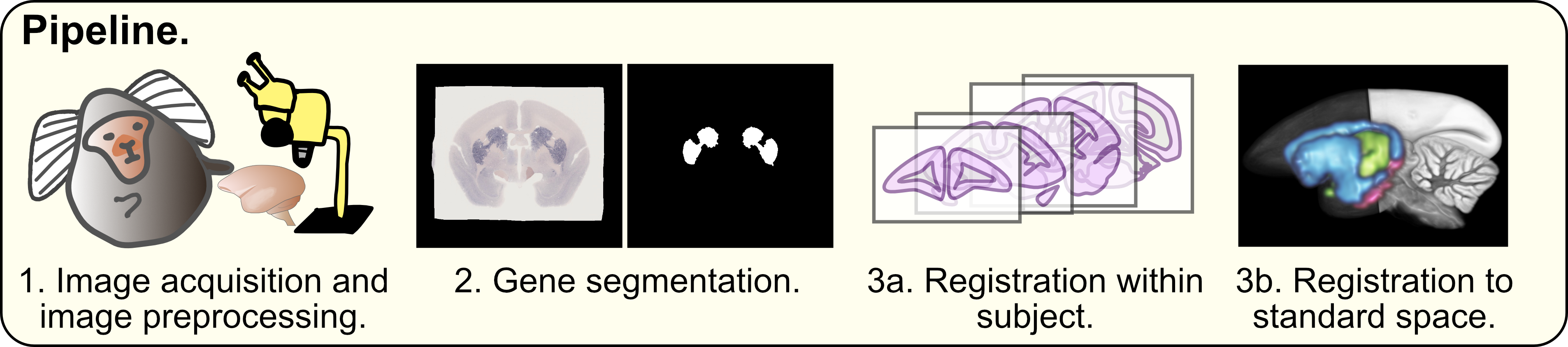}}
\caption{Schematic overview of the project pipeline to automate the creation of a gene atlas. Following image acquisition described in \cite{Shimogori2018}, images were registered and ISH gene expression was segmented from brain images.}
\label{fig:imgs/pipeline}
\end{figure}

Our goal is to develop an automated pipeline to create a gene expression atlas from ISH images, consisting of binary segmentations of gene expression from ISH images, registered to a standard space. We describe the image preprocessing, segmentation, and registration steps to achieve this for the adult marmoset brain (Figure \ref{fig:imgs/pipeline}). Segmentation of gene expression is necessary to clearly define areas of expression; true positive pixels are often difficult to discern in ISH images due to great variability in image contrast between images and in expression patterns between genes. We develop a semi-supervised deep learning segmentation model due to their superior performance over fully-supervised models in biomedical segmentation tasks despite fewer training labels \cite{Ibrahim2018}. Registration of ISH images is difficult to achieve because each gene has a unique expression pattern. Thus, we additionally develop an automated iterative algorithm that utilizes the Advanced Normalization Tools (ANTS) toolbox \cite{Avants2014} to register brain images to the Brain/MINDS Marmoset Connectivity Atlas (BMCA) template \cite{Skibbe2022}, to which neuronal tracer data, fiber tractography data, and anatomical labels have already been registered. Integration of the ISH dataset to the BMCA standard space will add transcriptomic data, facilitating a more holistic understanding of the marmoset brain. To our knowledge, this is the first report of automating the integration of marmoset ISH data into a standard space. Our code is publicly available: {\small\url{https://github.com/BrainImageAnalysis/MarmosetGeneAtlas_adult/MarmosetGeneAtlas_adult}}.

\section{Methodology}
\label{sec:format}

Data acquisition was conducted by the Laboratory for Molecular Mechanisms of Brain Development at the RIKEN Center for Brain Science \cite{Shimogori2018, Kita2021}. We describe the image analysis pipeline. 

\subsection{Preprocessing}

Data preprocessing consisted of downscaling, filtering, and morphological operations to remove artifacts. Metadata and data were reorganized to be in a machine-readable format.

\subsection{Segmentation}

To train the model, 3D image stacks of ISH gene expression from 14 genes (2470 2D images), were used in a 7:3 split for training and validation. To evaluate the model, 3D image stacks of ISH gene expression from five genes (520 2D images), which were separate from the training and validation datasets, were used. Ground truth segmentations were manually generated by an expert (CP).

\begin{figure}[htb]
\centering
\centerline{\includegraphics[width=8.5cm]{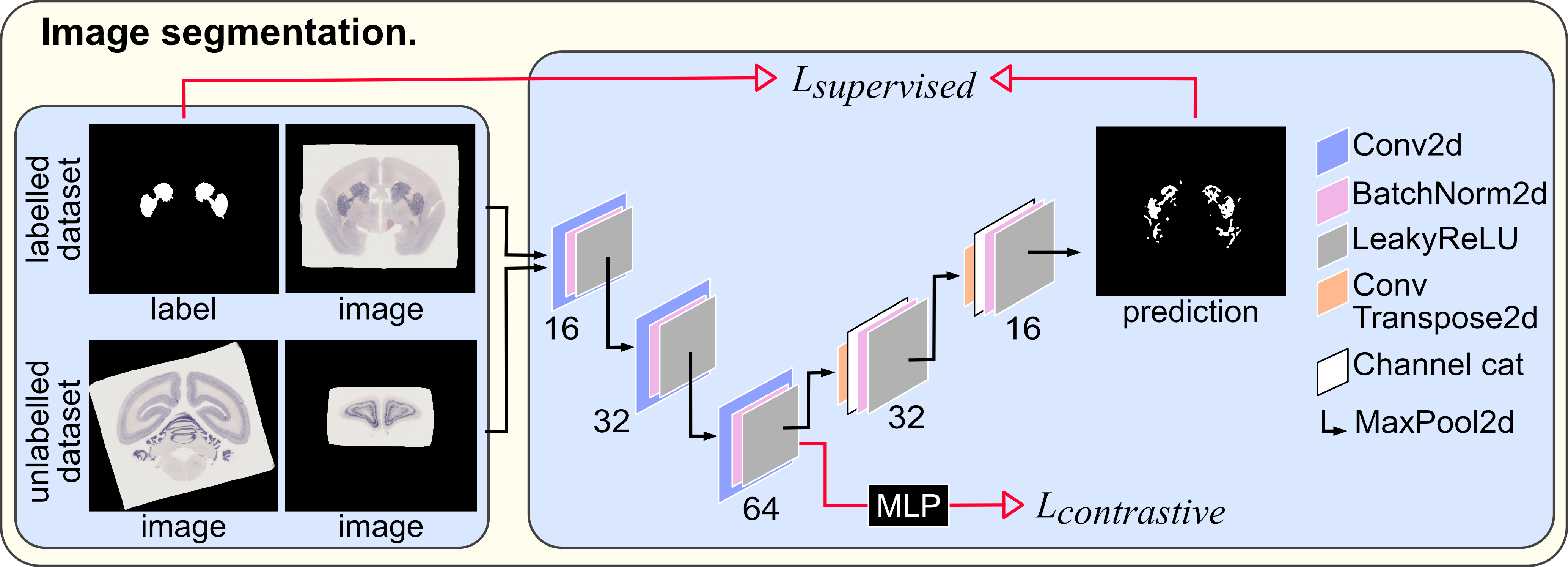}}
\caption{Segmentation model architecture. The semi-supervised segmentation model was based on a 2D U-Net, with supervised and contrastive losses.}
\label{fig:imgs/seg}
\end{figure}

The model was based on a 2D U-Net \cite{Ronneberger2015}, consisting of three levels (Figure \ref{fig:imgs/seg}). Each level in the encoder consisted of 2D convolution, batch normalization, and LeakyReLU layers. The number of features were doubled every step. In the decoder, 2D convolutions were replaced with 2D transposed convolutions. A sigmoid was applied to the output of the decoder. Input image patches were 400x400 pixels. 

The model was trained using the Adam optimization method and two losses, the supervised binary cross-entropy loss (\emph{L\textsubscript{supervised}}) and the unsupervised contrastive loss (\emph{L\textsubscript{contrastive}}).

The contrastive loss, previously described by Oord \textit{et al.} \cite{Oord2018} and Chen \textit{et al.} \cite{Chen2020}, shown in Equation \ref{eq:infonce}, calculates the loss between positive pairs of samples by maximizing agreement between features (\textit{z}) of two augmented views of the same image patch (positive pair: \textit{i,j}). In Equation \ref{eq:infonce}, $\tau$ is a temperature parameter and $\mathbb{I}_{k\not\equiv 1}$ is an indicator function. We used augmentations that were optimized by Chen \textit{et al.} \cite{Chen2020}: ColorJitter, RandomGrayscale, and GaussianBlur (Torchvision library). These augmentations vary the image contrast, brightness, hue, and saturation; parameters which already differ between images, and one reason why segmentation of this dataset difficult. The contrastive loss maximizes the agreement of image patches on the basis of image content, regardless of differences in colour profile and contrast. The contrastive loss was applied on features from the bottleneck layer of the model which were projected through a multilayer perceptron with one hidden layer (see \cite{Chen2020} for details). 

To train the model with both losses, skip connections were excluded to avoid leakage. We used a batch size of 16, which produced 30 negative samples for every positive pair. Code was written in PyTorch and PyTorch Lightning. Training was conducted using one NVIDIA A100 GPU.

\begin{equation}
l_{i,j}=-log\frac{exp(sim(z_{i},z_{j})/\tau)}{\sum_{k=1}^{2N}\mathbb{I}_{k\not\equiv 1}exp(sim(z_{i},z_{j})/\tau)}
\label{eq:infonce}
\end{equation}

We additionally trained a second version of the model, which was pretrained using the unsupervised contrastive loss only (\emph{model w/ pretraining} in \emph{Evaluation}). The dataset used for pretraining contained 186 unlabelled images.

\subsection{Registration}

\begin{figure}[htb]
\centering
\centerline{\includegraphics[width=\columnwidth]{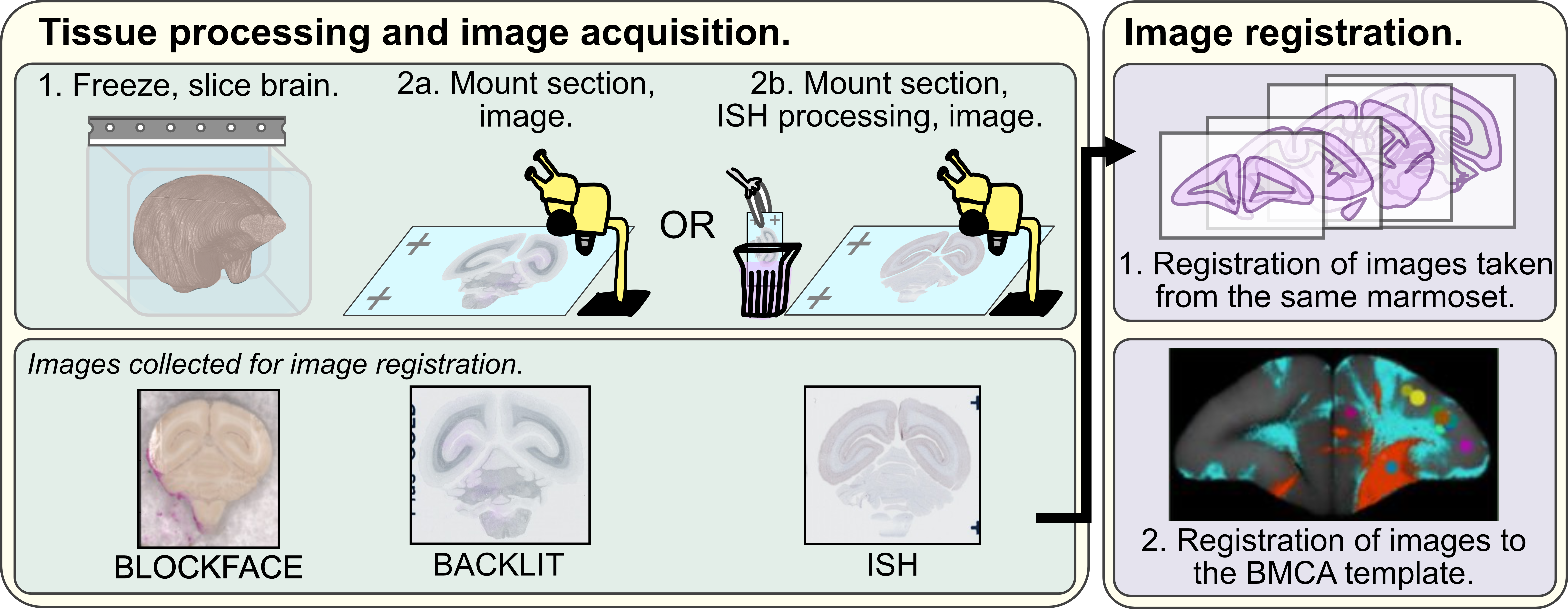}}
\caption{Tissue acquisition and image registration. During tissue acquisition, blockface, backlit, and ISH images were collected for image registration. ISH images were first registered within each subject and then to the BMCA template.}
\label{fig:imgs/reg}
\end{figure}

\begin{figure}[htb]
\centering
\centerline{\includegraphics[width=\columnwidth]{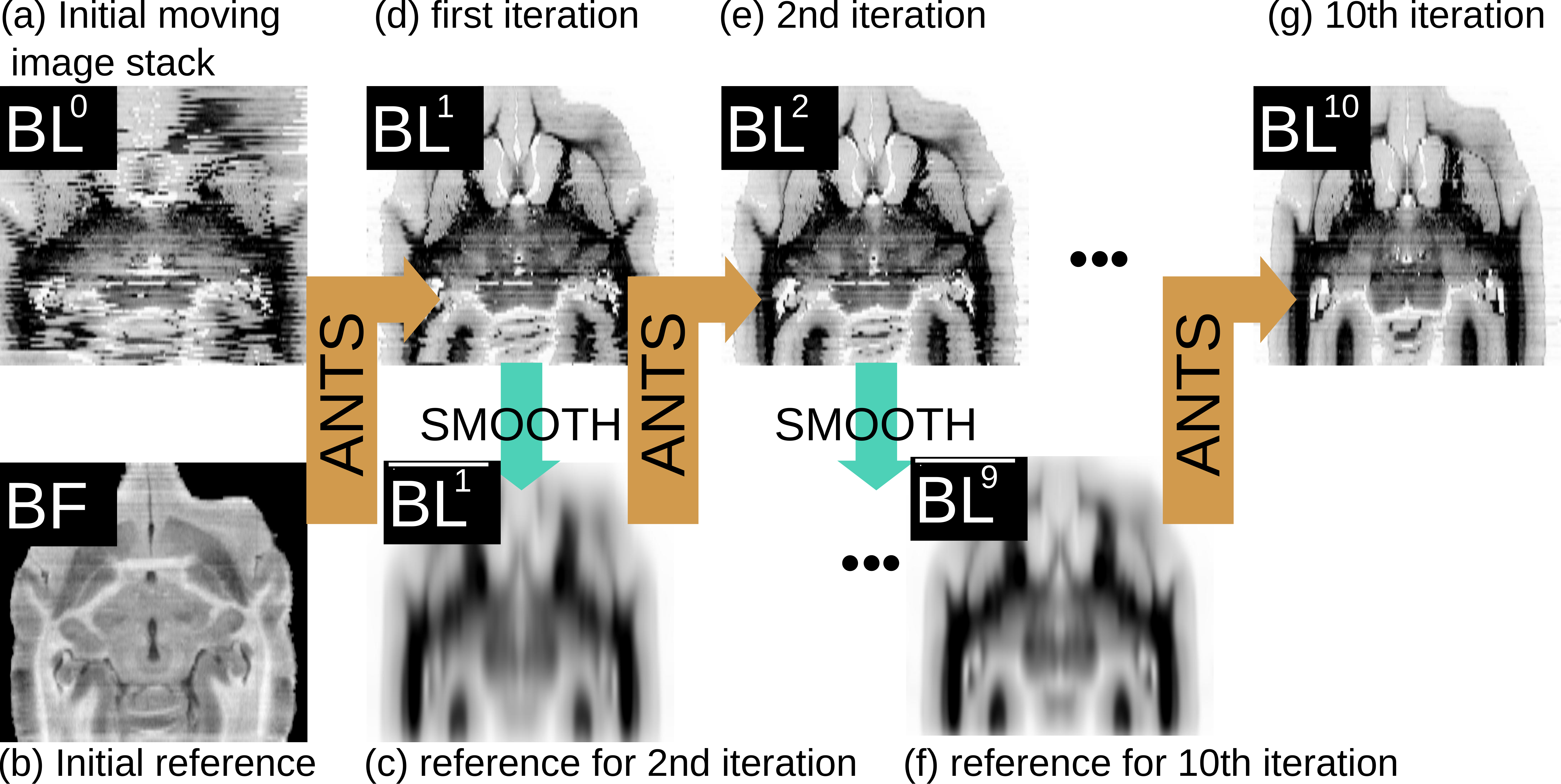}}
\caption{Iterative backlit registration.}
\label{fig:imgs/regpipe}
\end{figure}
%\vspace{-1.0cm}
We created an iterative algorithm using the ANTs Toolbox \cite{Avants2014} that creates a 3D brain image by automatically aligning and stacking the brain images to recover the original shape of the subject's brain, followed by registration to the BMCA reference marmoset brain template (Figure \ref{fig:imgs/reg}). 

To achieve the first stage of registration, blockface (\emph{BF}) and backlit (\emph{BL}) images were obtained during image acquisition (Figure \ref{fig:imgs/reg}). Blockface images are photos of the brain tissue before sectioning, and therefore show the shape of the brain with minimal spatial deformations. Backlit images are microscopy images of brain slices mounted onto slides (i.e. after sectioning), but prior to ISH or Nissl processing, and therefore are less spatially deformed than ISH and Nissl images. A reconstruction of the entire brain was achieved by concatenating the blockface images \cite{ABE2017102}. This blockface reconstruction was used as the first reference point to reconstruct a 3D image stack of backlit images (Figure \ref{fig:imgs/regpipe}). In the second step, we used the backlit image stack to align all ISH gene expression images. In the third step, we mapped all ISH gene expression image stacks to the BMCA template.
%Masks of blockface images, containing only brain tissue and excluding the ice block that brains were embedded in, were created using a 2D U-Net. 

For the reconstruction, we used an iterative algorithm that optimized the following objective function.

\begin{align}
%&\hat{T^{i}}=\underset{T}{\text{amin}}~a\mathcal{L}_a({BL^i_j} \circ T\circ %T^{k},\overline{{BL^{(i-1)}_j}})\nonumber\\
%&+b\mathcal{L}_b({BL^{(i-1)}_j} \circ T \circ T^{k},{BL^i_j})\nonumber\\
%&+c\mathcal{L}_c({BL^{i}_{j}} \circ T \circ T^{k},{BL^{(i-1)}_{(j-1)}})\nonumber\\
%&+c\mathcal{L}_c({BL^{i}_{j}} \circ T \circ T^{k},{BL^{(i-1)}_{(j+1)}}).
&\hat{T^{i}}=\underset{T}{\text{amin}}~a\mathcal{L}({BL^i_j} \circ T\circ T^{k},\overline{{BL^{(i-1)}_j}})\nonumber\\
&+b\mathcal{L}({BL^{(i-1)}_j} \circ T \circ T^{k},{BL^i_j})\nonumber\\
&+c\mathcal{L}({BL^{i}_{j}} \circ T \circ T^{k},{BL^{(i-1)}_{(j-1)}})\nonumber\\
&+c\mathcal{L}({BL^{i}_{j}} \circ T \circ T^{k},{BL^{(i-1)}_{(j+1)}}).
\label{eq:reg02}
\end{align}

\noindent Let $BL^i$ be the 3D backlit image stack after the $i$-th iteration, and let $BL^i_j$ the $j$-th 2D image section. $BF$ and $BF_j$ are the blockface image stack and its sections, respectively. Figure \ref{fig:imgs/regpipe} shows an overview of the backlit registration process. In the first step, ANTs's affine registration was used to map each 2D backlit section to its corresponding blockface section. The objective function is $\hat{T^0}=\underset{T}{\text{amin}}~\mathcal{L}({BL^0_j} \circ T,{BF^0_j})$, where $T$ is the objective, an affine transformation, and $\mathcal{L}$ is the normalized mutual information. After optimization, the process generated $\overline{BL^1}:=BL^1 * G(\sigma)$, a Gaussian smoothed 3D image of the aligned image stack, with $\sigma=3$ being the filter width. The next iterations used $\overline{BL^1}$ as the target image instead of the blockface image. In addition, we aimed for a smooth transition between neighbouring image sections. Therefore, three additional terms were added to the objective function; see Equation \eqref{eq:reg02}. The terms favor similarity with the previous iteration of the same section, but also with its predecessor and successor. We heuristically found that a=1, b=0.5, and c=0.5 worked best. $T^{k}$ is the transformation from a  previous iteration. Until iteration three, $T^{k}$ is the previous affine registration $T^{k}=T^{i-1}$. From iteration 3, we used deformable registration (SyN) instead of the affine registration, and set a=0, b=1, and c=0.25. $T^{k}=T^{2}$ was kept constant between iteration 3 and 6 to suppress high frequency artifacts from large non-linear deformations in the first SyN iterations. 

The next step was done separately for each gene. Each gene's images were registered to the newly created 3D backlit image stack. This was achieved in three steps. Since for each ISH section, there exists a corresponding backlit image, affine image registration was used to pre-align each ISH section to its corresponding backlit counterpart. Two additional SyN iterations were used to reconstruct the ISH 3D image stacks. The loss function in the last two iterations was similar to \eqref{eq:reg02}, where  $\overline{{BL^{(i-1)}_j}}$, ${BL^{(i-1)}_{(j-1)}}$ and ${BL^{(i-1)}_{(j+1)}}$ were replaced with their ISH counterparts. 

In the final step, a 3D affine and 3D SyN registration were applied to map the 3D backlit image, and therefore the ISH images, to the BMCA 3D marmoset brain reference space.

%$\overline{BL^i}:=BL^i * G(\sigma)$

%\begin{align}
%\hat{T^0}=\underset{T}{\text{amin}}~\mathcal{L}({BL^0_j} \circ T,{BF^0_j})
%\label{eq:reg01}
%\end{align}

\section{Evaluation}
\label{sec:evaluation}

To evaluate the segmentation model, model outputs with and without pretraining (\emph{model w/ pretraining} and \emph{model}) were compared to ground truth segmentations (\emph{gt}), two other human-generated sources (\emph{thresholded}, \emph{manual}), and one other machine-generated source (\emph{unet}), summarized below:

\begin{itemize}
  \item \emph{gt}: ground-truth, manually generated by CP
  \item \emph{thresholded}: thresholded images, thresholds were manually set for each image by CP
  \item \emph{manual}: manually generated by five other annotators (MFR, MB, MS, BX, HS) to evaluate the consistency among human annotators
  \item \emph{model}: our model without pretraining
  \item \emph{model w/ pretrain}: our model with pretraining
  \item \emph{unet}: fully-supervised vanilla 2D three-level UNet
\end{itemize}

Quantitatively, segmentations were evaluated using the Dice score; we report the mean and standard deviation in Table \ref{tab:quanttable}. Our model outperformed all other methods by a wide margin. High standard deviations observed in human-generated segmentations (\emph{thresholded*} and \emph{manual*}), and overall low Dice scores ($<$0.5) show the difficulty in segmenting gene expression from ISH images due to variations in expression patterns between genes and differences in image contrast even for images obtained from the same marmoset. High standard deviation observed in \emph{model w/ pretraining} segmentations can likely be improved with longer pretraining and optimization of augmentations. 

% version of the table without 'vs gt'
\begin{comment}
\begin{table}[h]
\centering
\begin{tabular}{@{}ccc@{}}
\toprule
{\color[HTML]{2E3436} }                     & {\color[HTML]{2E3436} Dice (mean)} & {\color[HTML]{2E3436} Dice (SD)} \\ \midrule
{\color[HTML]{2E3436} thresholded*} & {\color[HTML]{2E3436} 0.3629} & {\color[HTML]{2E3436} 0.2981} \\
{\color[HTML]{2E3436} manual*}              & {\color[HTML]{2E3436} 0.1815}      & {\color[HTML]{2E3436} 0.2965}    \\
{\color[HTML]{2E3436} model}                & {\color[HTML]{2E3436} 0.4948}      & 0.2512                           \\
{\color[HTML]{2E3436} model w/ pretraining} & {\color[HTML]{2E3436} 0.4050}      & 0.2996                           \\
{\color[HTML]{2E3436} unet}                 & {\color[HTML]{2E3436} 0.2581}      & 0.2124                           \\ \bottomrule
\end{tabular}
\end{table}
\end{comment}

% version of the table with 'vs gt'
\begin{table}[ht]
\centering
\begin{tabular}{@{}ccc@{}}
\toprule
{\color[HTML]{2E3436} }                               & {\color[HTML]{2E3436} Dice (mean)} & {\color[HTML]{2E3436} Dice (SD)} \\ \midrule
{\color[HTML]{2E3436} thresholded* vs gt*} & {\color[HTML]{2E3436} 0.3629} & {\color[HTML]{2E3436} 0.2981} \\
{\color[HTML]{2E3436} manual* vs gt*}      & {\color[HTML]{2E3436} 0.1815} & {\color[HTML]{2E3436} 0.2965} \\
{\color[HTML]{2E3436} model vs gt*}                & {\color[HTML]{2E3436} 0.4948}      & 0.2512                           \\
{\color[HTML]{2E3436} model w/ pretraining vs gt*} & {\color[HTML]{2E3436} 0.4050}      & 0.2996                           \\
{\color[HTML]{2E3436} unet vs gt*}                 & {\color[HTML]{2E3436} 0.2581}      & 0.2124                           \\ \bottomrule
\end{tabular}
\caption{\label{tab:quanttable}Quantitative evaluation of segmentations. Human-generated segmentations are marked by a *.}
\end{table}

\begin{figure}[htb]
\centering
\centerline{\includegraphics[width=8.5cm]{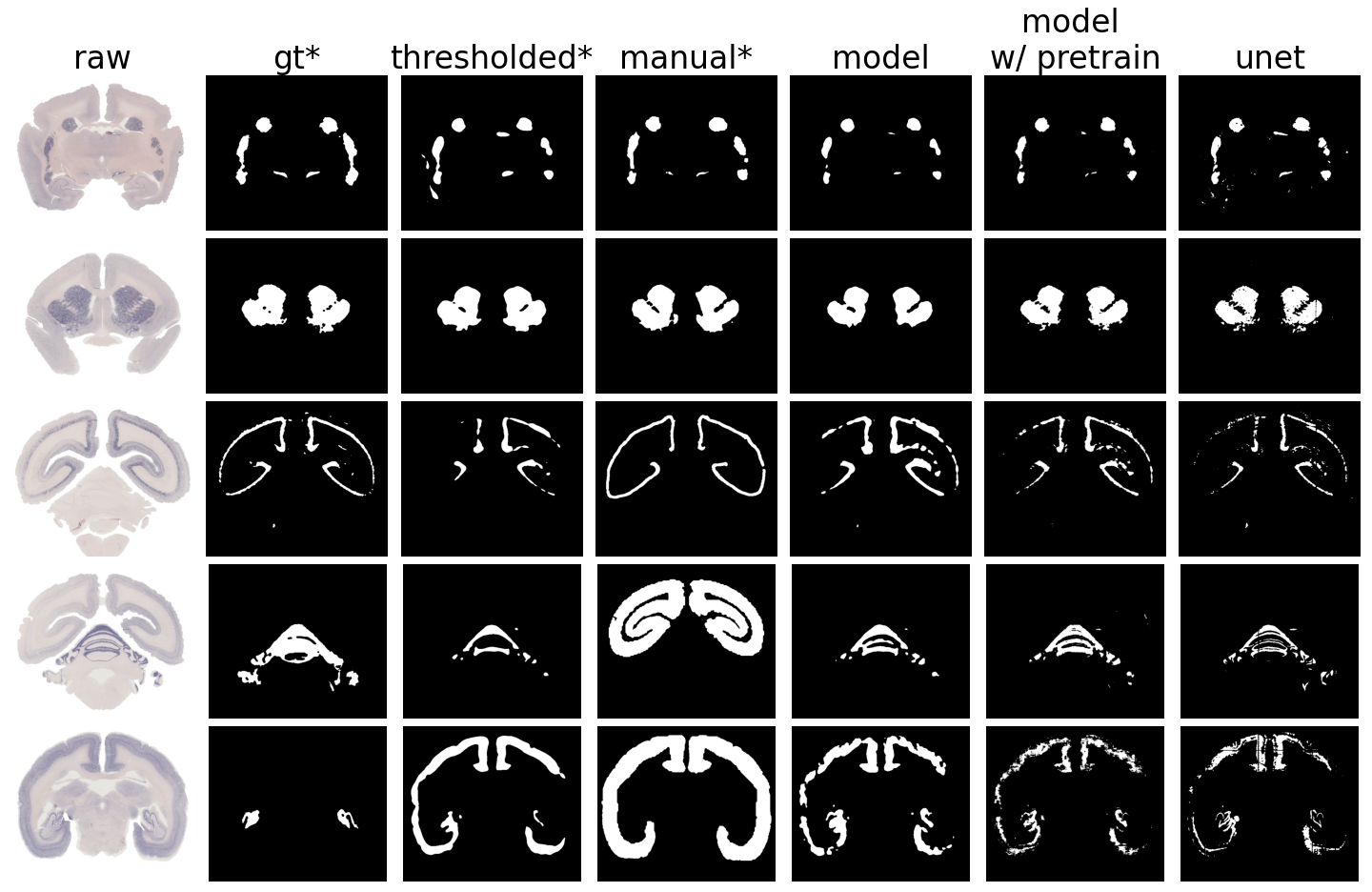}}
\caption{Qualitative comparison of segmentations. Human-generated segmentations are marked by *. The challenge of gene expression segmentation is exemplified in row 5, where all methods segmented the wrong structure (see \emph{gt} for the correct structure).}
\label{fig:imgs/qual}
\end{figure}

A sample of segmentations are shown in Figure \ref{fig:imgs/qual}. Qualitatively, it can be seen \emph{manual*} segmentations performed the worst (see row 4, where all methods segmented the correct structure except for \emph{manual*}).

\subsection{Automated stack alignment}
\label{sec:stackalgn}

\begin{figure}[htb]
\centering
\centerline{\includegraphics[width=\columnwidth]{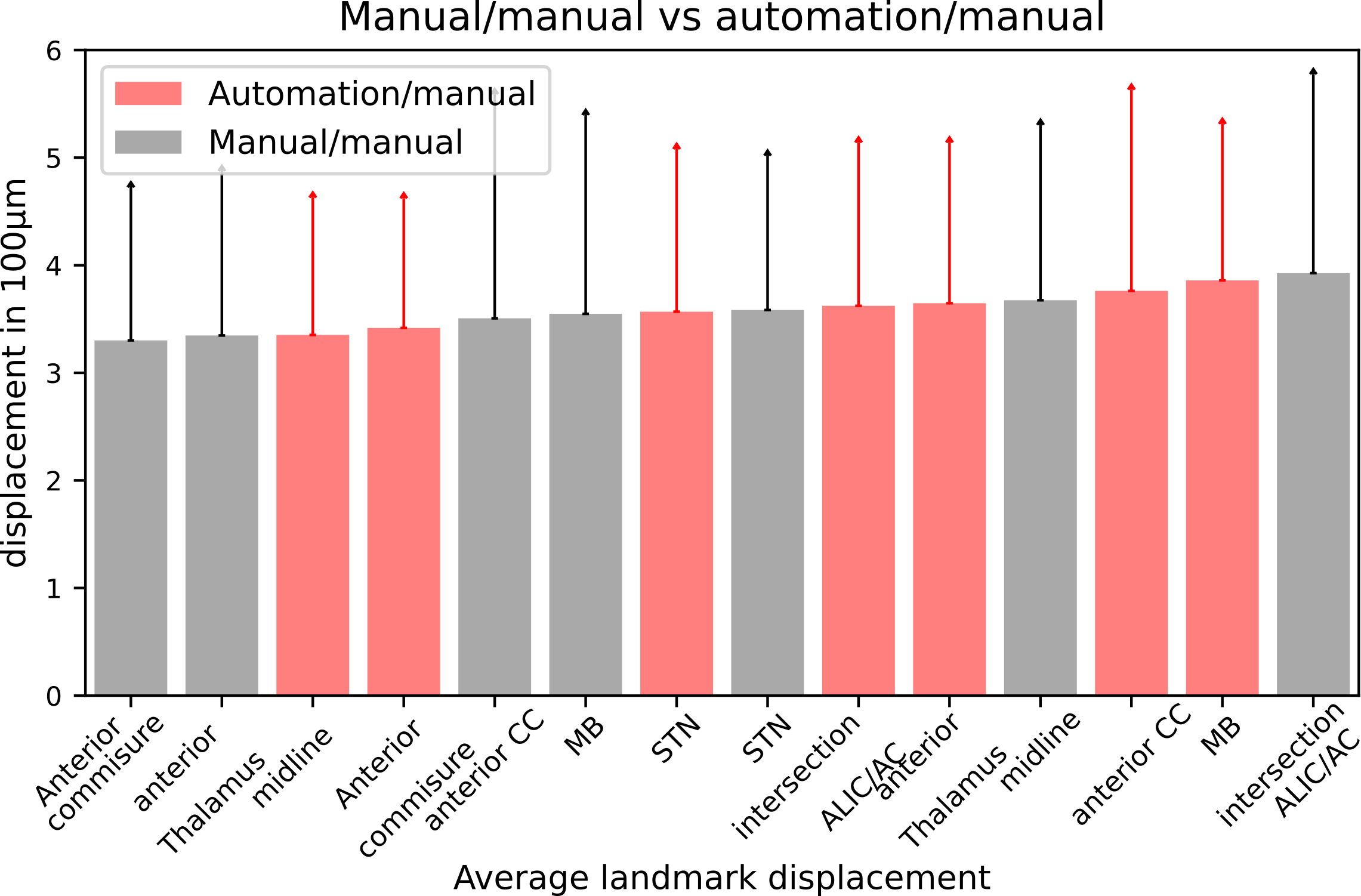}}
\caption{Automated stack alignment could maintain the accuracy of manually placed landmarks in the adult marmoset brain.}
\label{fig:imgs/regeval}
\end{figure}

To assess the quality of 3D stack alignment, we defined seven landmarks in the reference template of the marmoset brain: (single points unless indicated otherwise): anterior commissure, anterior thalamus, midline, dorsal tip of the anterior cingulate cortex (\emph{CC}), posterior commissure of the midbrain (\emph{MB}, two points), subthalamic nucleus (\emph{STN}, two points), and the intersection of the anterior limb of the internal capsule and the anterior commissure (\emph{intersection ALIC/AC}, two points). For each ISH image stack, three experts manually placed the landmarks. For comparison, landmarks were automatically mapped based on the transformation fields generated by the image registration pipeline. The smaller the displacement between a pair of landmarks manually placed by two different annotators, the better the agreement. The same comparison was done between manual landmarks and automatically mapped landmarks. The median match between manual annotations and automation was compared to the best match between two human annotations, which gave an advantage to human annotations. Figure \ref{fig:imgs/regeval} shows the scores sorted by displacement, shown in units of 100 $\mu$m. In this scenario, automation could maintain the performance of manual methods. Of note, if we took the median displacement between manually placed landmarks as well, automation outperformed manual methods for all landmarks. 

\section{Conclusion}
We describe the novel development of an automated pipeline to integrate adult marmoset gene expression data into a standard space. Quantitative and qualitative evaluations showed that the unsupervised contrastive loss improved segmentation of ISH gene expression. We expect that pretraining with a greater number of unlabelled images and optimizing augmentation parameters for the ISH dataset will improve performance. High standard deviation in human-generated segmentations show the unreliability of manual labelling. Comparison of registration annotations between automation and manual methods revealed that automation also performed on par with humans. We plan to explore deep learning registration methods to improve registration \cite{Hoopes2021, Balakrishnan2019}, as well as other segmentation models. This automated pipeline can be used to process and integrate data from different imaging modalities for co-visualization and comparative analyses.

\section{Compliance with ethical standards}
\label{sec:compliance}
This research study was conducted retrospectively using marmoset imaging data made available in open access at https://gene-atlas.brainminds.riken.jp/. The use of marmosets followed the guidelines of and were approved by the RIKEN Institutional Animal Care Committee, described in \cite{Shimogori2018, Kita2021}.

\section{Acknowledgments}
\label{sec:acknowledgments}
This work was supported by the Japan AMED (JP15dm0207001) and the Japan Society for the Promotion of Science. All authors state no potential conflicts of interest.
This work has been submitted to the IEEE for possible publication. Copyright may be transferred without notice, after which this version may no longer be accessible.

\bibliographystyle{IEEEbib}
%\bibliography{strings,refs}
\bibliography{refs}

\end{document}